\documentclass[runningheads]{llncs}
\usepackage{comment}
\usepackage{graphicx}
\usepackage{amsmath}
\usepackage{amsfonts}
\usepackage{appendix}
\usepackage{booktabs}
\usepackage{caption}
\begin{document}
\title{Disentangled Variational Information Bottleneck for Multiview Representation Learning}



\titlerunning{Disentangled Variational Information Bottleneck}
%
\author{Feng Bao}
\authorrunning{Feng Bao}
%
\institute{University of California, San Francisco\\\email{feng.bao@ucsf.edu} }

\maketitle              

\begin{abstract}

Multiview data contain information from multiple modalities and have potentials to provide more comprehensive features for diverse machine learning tasks. A fundamental question in multiview analysis is what is the additional information brought by additional views and can quantitatively identify this additional information. In this work, we try to tackle this challenge by decomposing the entangled multiview features into shared latent representations that are common across all views and private representations that are specific to each single view. We formulate this feature disentanglement in the framework of information bottleneck and propose disentangled variational information bottleneck (DVIB). DVIB explicitly  defines the properties of shared and private representations using constrains from mutual information. By deriving variational upper and lower bounds of mutual information terms, representations are efficiently optimized. We demonstrate 
the shared and private representations learned by DVIB well preserve the common labels shared between two views and  unique labels corresponding to each single view, respectively. DVIB also shows comparable performance in classification task on images with corruptions. DVIB implementation is available at \url{https://github.com/feng-bao-ucsf/DVIB}.

\end{abstract}
\section{Introduction}

With advances in the past decade, performances of major machine learning frameworks have reached their accuracy plateau in many tasks \cite{beyer2020we,tsipras2020imagenet,yoshida2019data}. To further overcome the performance bottleneck, multiview learning methods are viewed as promising solutions \cite{xu2013survey,wang2015deep}. By collecting additional views from samples, we expect to obtain more useful and task-relevant features, therefore enhancing the performance of methods through increasing the information abundance within the data \cite{DBLP:conf/icml/NgiamKKNLN11,federici2020learning,li2018survey}.

In multiview data, each modality is collected using different technologies and approaches and contains different levels of  corruptions, noises and/or missings. One fundamental and critical question in multiview  analysis is: can additional views provide additional effective information to facilitate the learning tasks compared with single view data? If yes, can we explicitly identify the additional information to explain the view property and enhance the data interpretability?

\begin{figure}
\centering
\includegraphics[width=0.8\textwidth]{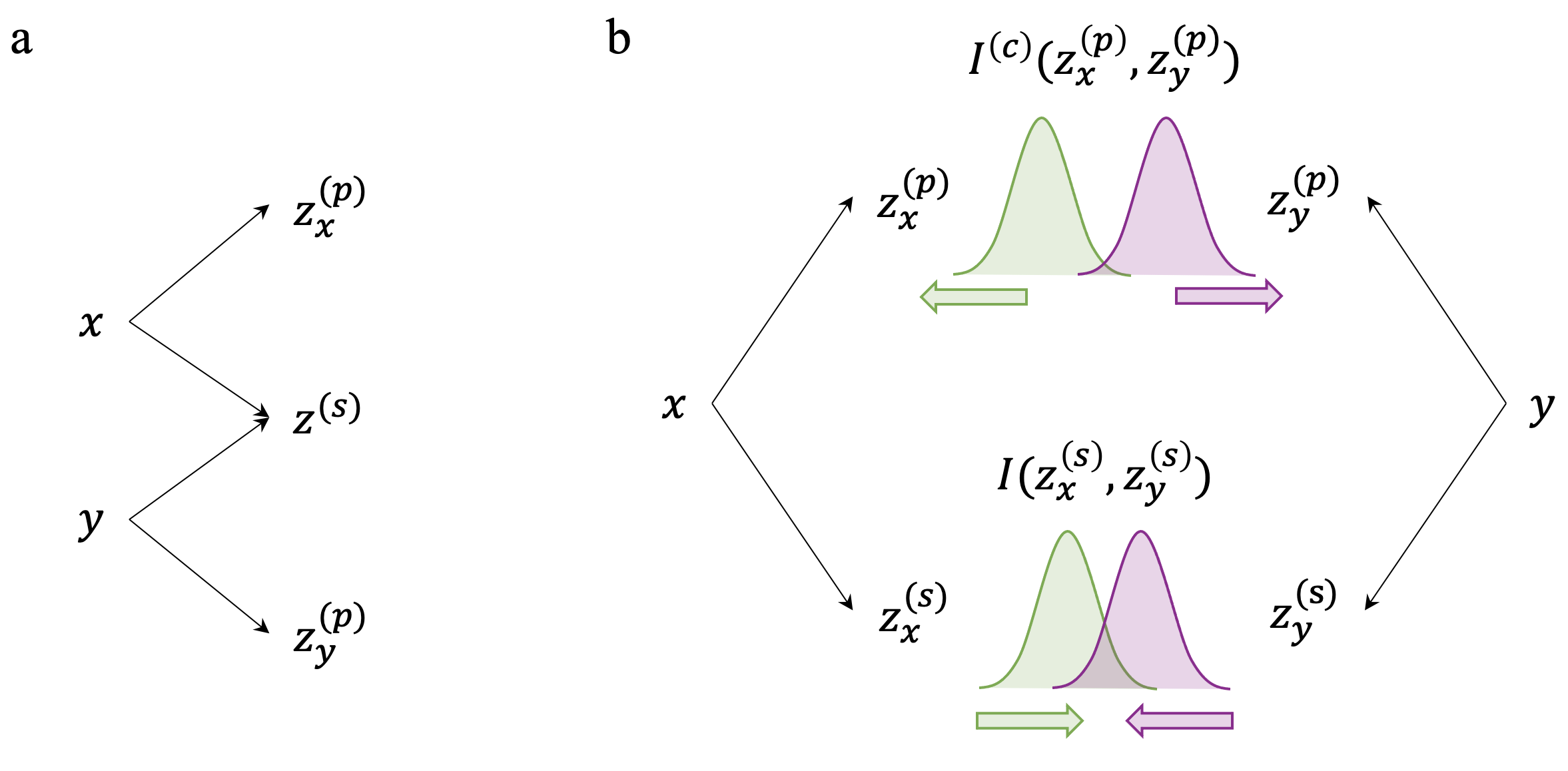}
\caption{Illustrations of (a) multiview disentangled representation learning concept and (b) proposed  disentangle variational information bottleneck (DVIB) method. } \label{intro}
\end{figure}

To answer these questions, it requires us to decompose the entangled information embedded in multi-view data into view-shared and view-private (a.k.a view-specific) representations \cite{wang2016deep,higgins2016beta} (Fig. \ref{intro}a). Based on the view decomposition, contributions from each single view can be explicitly quantified and analyzed. Besides that, the view-shared information exhibits the general and common features of the sample and can be used to reduce the effects of data corruption and noise \cite{wang2016deep,andrew2013deep}. Meanwhile view-private information represents the unique properties from single modality therefore can be used to evaluate its importance to specific tasks and reflect the strength and weakness of technologies that generate the view.

Learning disentangled representation from multiview data is challenging in terms of the modeling of the entanglement \cite{federici2020learning,wang2016deep,higgins2016beta}. In this work, we formulate the disentangled representation learning in the framework of  information bottleneck and propose disentangled variational information bottleneck (DVIB). In the  optimization target (see Section \ref{sec_method}), we aim to maximize the mutual information between shared latent representations generated from different views while minimizing the mutual information between private representations at the same time (Fig. \ref{intro}b). With such constrains, the properties of  private and shared representations are  explicitly formulated. The learning target can be efficiently optimized through deriving variational bounds and auxiliary cost functions. 
We demonstrate the ability of DVIB in accurately decomposing the common or view-specific information from multiview data and improving the robustness in classification task on large-scale datasets.

\section{Preliminaries and existing works}

Representation learning is to extract effective, highly compact features from raw data containing various levels of noises and corruptions \cite{li2018survey,xu2013survey,bengio2013representation}. Efficient design of  representation methods can greatly facilitate the down-streaming learning tasks. Learning highly compact representations from the view of information theory has attracted long-time attentions \cite{deng2016deep,tishby2015deep}. Pioneering work information bottleneck \cite{tishby2000information,amjad2019learning}
aims to learn the representation from minimal input information but can predict the outcome well, by maximizing mutual information between the latent representations and output while minimizing mutual information between the input and latents. 

The information bottleneck defines an elegant target for the optimization of compact representations from the view of information theory. 
However, efficiently calculation of the mutual information is challenging due to the intractable estimation of marginal distributions \cite{belghazi2018mutual,poole2019variational}. Recently, approximation of mutual information has been advanced greatly with the help of variational inference \cite{poole2019variational}. By replacing intractable margins with tractable approximators, we can alternatively seek to derive the variational upper or lower bound of mutual information. With the variational neural network and re-parameterization tricks \cite{kingma2013auto}, we can efficiently optimize the mutual information bounds.

The combination of information bottleneck and variational inference leads to the development of variational information bottleneck (VIB) \cite{alemi2016deep}. VIB is able to learn maximally informative representations and shows robust performance with existence of perturbations. Following the idea of VIB, recently proposed  methods enable more flexible representation learning for classification task \cite{hjelm2018learning} and information harmonization between multiview data (MVIB) \cite{federici2020learning}. 

Learning disentangled representation is an attracting task \cite{tran2017disentangled,zhang2019gait}.  Some recent works, e.g. ($\beta$-VAE) \cite{higgins2016beta} try to formulate general disentangled information learning  from the view of variational inference and it is proved to be closely related to the information bottleneck \cite{burgess2018understanding}.  However, explicitly modeling and quantifying the disentanglement is a challenge. The recent development of mutual information and information bottleneck methods provides a theoretical fundamental of our method.

\section{Disentangled variational information bottleneck}
\label{sec_method}

To simplify the description, here we consider two views of features $x$ and
$y$ collected from the same sample. Our goal is to decompose each single feature
view, for example $x$, into two latent representations $z_x^{(s)}$ and
$z_x^{(p)}$, where $z_x^{(s)}$ is the shared representation that  preserves the common information from both views, and $z_x^{(p)}$ is the private representation that exhibits the view-specific property in $x$. Similarly, we can define the shared and private latents as $z_y^{(s)}$ and
$z_y^{(p)}$ for view $y$.

\subsection{Information bottleneck for the shared representation}

For the shared representations, we expect to capture the information shared by both $x$ and $y$ meanwhile neglecting the view-specific information. 
Following the definition of information bottleneck, we formulate the learning targets of $z_x^{(s)}$ and $z_y^{(s)}$ using mutual information,

\begin{equation}
\max{~I(x; z_x^{(s)})+\lambda_xI(y; z_x^{(s)})}\label{eq1}
\end{equation}

\begin{equation}
\max{~I(y; z_y^{(s)})+\lambda_yI(x; z_y^{(s)})}\label{eq2}
\end{equation}
where in Eqs.\ref{eq1} and \ref{eq2}, first terms require the shared representation $z_*^{(s)}$ to have maximal mutual information with the view where it was generated from. The second term forces the shared representation, even it was learned from one view, can maintain high mutual information with the other view. Hyperparameters $\lambda_x\ge0$ and $\lambda_y\ge0$ balance the relative importance of two mutual informations.
By maximizing two target functions, we can explicitly constrain shared latent representations to maintain maximal mutual information for both views simultaneously. 

\subsection{Information bottleneck for the private representation}

For the private representations, we restrict the learned $z_x^{(p)}$ and $z_y^{(p)}$ to only contain the unique information from the view it is generated from, but no information form the other view. Similarly, we define the learning targets as:

\begin{equation}
\max{~I(x; z_x^{(p)})-\beta_xI(y; z_x^{(p)})}\label{eq3}
\end{equation}

\begin{equation}
\max{~I(y; z_y^{(p)})-\beta_xI(x; z_y^{(p)})}\label{eq4}
\end{equation}

Similar as the information bottleneck \cite{tishby2000information}, the maximization of first terms in Eqs. \ref{eq3} and \ref{eq4} require private latents and raw features ($z_x^{(p)}$ and $x$, or $z_y^{(p)}$ and $y$) to stay as similar as possible by mutual information metric.
Meanwhile, the second terms require the $z_x^{(p)}$ (resp. $z_y^{(p)}$ ) to contain as less information as possible form the view $y$ (resp. $x$). Again, non-negative hyperparameters $\beta_x$ and $\beta_y$ define the trade-offs between the information to gain from the view where the private representation is learned and the information to suppress from the other view.







\subsection{Variational bounds}

The optimization of introduced targets  requires the calculation of a number of mutual information terms. However, it is known the mutual information is intractable for high dimensional variables \cite{belghazi2018mutual,kingma2013auto,alemi2016deep}. We alternatively sort to derive the variational bounds of mutual information \cite{poole2019variational}. 

\subsubsection{Lower bound of $I(x,z_x^{(s)})$}

We first consider the lower bound of mutual information between the latent representation (either shared or private) and the view it was generated from. Taking $I(x,z_x^{(s)})$ as an example, we have

\begin{align}  
  \begin{aligned}[t]
  I(x,z_x^{(s)})  &= \mathbb{E}_{p(x,z_x^{(s)})}\log{\frac{p(x|z_x^{(s)})}{p(x)}}\\
              &= \mathbb{E}_{p(x,z_x^{(s)})}\log{\frac{p(x|z_x^{(s)})}{q(x|z_x^{(s)})}\frac{q(x|z_x^{(s)})}{p(x)}}\\
              &= \mathbb{E}_{p(x,z_x^{(s)})}\log{\frac{q(x|z_x^{(s)})}{p(x)}}+\mathbb{E}_{p(z_x^{(s)})}KL[p(x|z_x^{(s)})||q(x|z_x^{(s)})]\\
              &\ge \mathbb{E}_{p(x,z_x^{(s)})}\log{q(x|z_x^{(s)})} + H(X)
  \end{aligned}
\end{align}
where $KL[p||q]\ge 0$ represents the Kullback–Leibler divergence of two variables $p$ and $q$; $H(X)$ is the entropy of $x$; $q(x|z_x^{(s)})$ is the variational approximation of conditional distribution $p(x|z_x^{(s)})$. 
Entropy term $H(X)$ is determined by the dataset and is independent of the optimization process. 
Therefore, to maximize the $I(x,z_x^{(s)})$, we can alternatively maximize the $\mathbb{E}_{p(x,z_x^{(s)})}q(x|z_x^{(s)})$.
Similarly, we can derive the lower bounds for $I(x,z_x^{(p)})$, $I(y,z_y^{(p)})$ and $I(y,z_y^{(s)})$. The full derivation of variational bounds can be found in Appendix \ref{app1}.


\subsubsection{Lower bound of $I(y,z_x^{(s)})$}

Next we consider the mutual information between shared latent representations and features that are from different views. We take $I(y,z_x^{(s)})$ as an example. We follow the derivation in Ref.\cite{federici2020learning} and write the lower bound as 

\begin{align}  
  \begin{aligned}[t]\label{prior_x}
  I(y,z_x^{(s)}) 	&= I(z_x^{(s)},z_y^{(s)})+I(z_x^{(s)};y|z_y^{(s)})\\
  &\ge I(z_x^{(s)},z_y^{(s)})\\
  \end{aligned}
\end{align}
where the mutual information between a raw feature and a latent variable is approximated by the mutual information between two shared latent representations $z_x^{(s)}$ and $z_y^{(s)}$. 
We can derive the lower bound for $I(x,z_y^{(s)})$ symmetrically.
Appendix \ref{app2} and Appendix Fig. 1 provide the derivation and illustration of the lower bound. Therefore, we can combine the optimization of $I(x,z_y^{(s)})$ and $I(y,z_x^{(s)})$ to the same target $ I(z_y^{(s)},z_x^{(s)})$.

\subsubsection{Upper bound of $I(y;z_x^{(p)})$}

Finally we consider the view-private latent representations $z_x^{(p)}$ and $z_y^{(p)}$. Again, we take the mutual information $I(y;z_x^{(p)})$ as an example and derive the upper bound. We have:


\begin{align}
  \begin{aligned}[t]\label{eq_up}
  I(y,z_x^{(p)})  &= \mathbb{E}_{p(y,z_x^{(p)})}\log{\frac{p_x^{(p)}(z_x^{(p)}|y)}{p(z_x^{(p)})}}\\
  &= \mathbb{E}_{p(y,z_x^{(p)})}\log{\frac{p_x^{(p)}(z_x^{(p)}|y)}{r(z_x^{(p)})}\frac{r(z_x^{(p)})}{p(z_x^{(p)})}}\\
  &= \mathbb{E}_{p(y,z_x^{(p)})}\log{\frac{p_x^{(p)}(z_x^{(p)}|y)}{r(z_x^{(p)})}}-KL[r(z_x^{(p)})||p(z_x^{(p)})]\\
  &\le \mathbb{E}_{p(y,z_x^{(p)})}\log{\frac{p_x^{(p)}(z_x^{(p)}|y)}{r(z_x^{(p)})}}\\
  &= \mathbb{E}_{p(y,z_x^{(p)})}\log{\frac{p_x^{(p)}(z_x^{(p)}|y)}{p_y^{(p)}(z_y^{(p)}|y)}}+\mathbb{E}_{p(y,z_x^{(p)})}\log{\frac{p_y^{(p)}(z_y^{(p)}|y)}{r(z_x^{(p)})}}\\
  \end{aligned}
\end{align}
where $p_x^{(p)}$ and $p_y^{(p)}$ represent private encoders that learn private latent representations $z_x^{(p)}, z_y^{(p)}$ from raw features $x, y$.
The upper bound tights on the approximation of marginal distribution $r(z_x^{(p)})$ to prior $p(z_x^{(p)})$. Two terms are in the upper bound: the first term is the encoding difference of two latent representations from $p_x^{(p)}$ and $p_y^{(p)}$ but with the same input $y$. It encourages two encoders to produce inconsistent encoding. The second term is the difference between encoder $p_y^{(p)}$ with the approximated margin $r(z_x^{(p)})$. Minimizing the upper bound requires the encoder from $y$ generates representations that are heterogeneous with both posterior and prior of $z_x^{(p)}$.
Appendix \ref{app3} provides complete derivation.

Here, the estimation of the first term is not easy as it requires to input $y$ to encoder $p_x^{(p)}$. Because we use stochastic neural network mapping from raw data to the latent, the outputs from two view encoders can be greatly different. Besides, the second term enlarges the differences between  $z_x^{(p)}$ prior and view-$y$ private encoder output and has the same optimization direction as the first term. Therefore, we simplify the upper bound to 

\begin{align}
  \begin{aligned}[t]
  \min{I(y,z_x^{(p)})}  &\equiv  \min{\mathbb{E}_{p(y,z_x^{(p)})}\log{\frac{p_y^{(p)}(z_y^{(p)}|y)}{r(z_x^{(p)})}}}\\
  \end{aligned}
\end{align}


We note this formulation is the same as negative mutual information term in variational  information bottleneck \cite{alemi2016deep}, which constrains information flow from  raw features to the latent representations. Therefore, our optimization target has the same function to control the information flow  from two views to latents.

\subsection{Optimization}

Finally, we combine the bounds and auxiliary targets introduced in the previous section together and derive the overall optimization function:

\begin{align}  
  \begin{aligned}[t]
  \max ~I_{total}&=I_x+I_y\\ 
  &\ge I_{LB}(x;z_x^{(s)};z_x^{(p)})+ I_{LB}(y;z_y^{(s)};z_y^{(p)}) + \\
  &~~~~\lambda I(z_x^{(s)},z_y^{(s)})  - \beta I^{(c)}(z_x^{(p)},z_y^{(p)})\\
  \end{aligned}
\end{align}
where the full formulation of  each term is given in Appendix \ref{app4}.  $I_{LB}(x;z_x^{(s)};z_x^{(p)})$ and $I_{LB}(y;z_y^{(s)};z_y^{(p)})$ are lower bounds of $I(x;z_x^{(s)};z_x^{(p)})$ and $I(y;z_y^{(s)};z_y^{(p)})$, respectively; $I^{(c)}(z_x^{(p)},z_y^{(p)})$ represents the cross mutual information between private representations of two views (Appendix \ref{app4}). 

To optimize the target function, we make use of the variational autoencoder structure and employ four encoders to output parameters that define posteriors of $z_x^{(s)}$, $z_y^{(s)}$, $z_x^{(p)}$ and $z_y^{(p)}$ while decoders are used to reconstruct raw features from latent representations. To maximize $I(z_x^{(s)},z_y^{(s)})$, we use the neural mutual information estimators \cite{amjad2019learning,poole2019variational,belghazi2018mine}. The exceptions over joint distributions are approximated by the empirically joint distributions \cite{federici2020learning,alemi2016deep}.

\section{Experiments}

In this section, we calibrate the performance of DVIB and evaluate the quality of shared and private representations using various datasets. In the experiments, we focus on two questions: 
1) the ability of private and shared latent representations to decompose entangled information and capture meaningful contents from each view;
2) how can the representations facilitate down-streaming analysis.

We implemented the DVIB in the framework of variational autoencoder where encoder and decoder networks (simple multi-layer neural network) were used to learn the representation distribution and reconstruct original signal. For prior distributions $r(z_x^{(p)})$ and $r(z_y^{(p)})$, we restrict them to follow $N(0,I)$ as described in\cite{alemi2016deep}. To estimate the mutual information $I(z_x^{(s)},z_y^{(s)})$, our implementation employed the Jensen-Shannon estimator \cite{federici2020learning,belghazi2018mutual}.

\subsection{Evaluation of information disentanglement on MNIST}

We start from MNIST handwriting digit dataset \cite{deng2012mnist}, which involves simple sample categories and is easy to generate paired multiview data through image transformations. This can be an example to set up the baseline performance of DVIB.
Here, for every sample in MNIST, we consider two transformation to simulate two-view data: 1) a rotation of the image in one of the following angle $[0, \pi/16,\pi/8, 3\pi/16,\pi/4]$ to generate view $x$; 2) a random flip from [None, horizontal, vertical, horizontal + vertical] to generate view $y$.
We note each transformation choice is randomly performed on each digit so that the transformation is independent of the original MNIST digit labels. Because of the simple network used in DVIB cannot efficiently capture complicate image transformation, we firstly feed two view data into Inception-v3\footnote{Pretrained model provided by TensorFlow Hub (https://tfhub.dev/).} that was pretrained on ImageNet \cite{deng2009imagenet} and the output of last fully connected layers (2,048 dimensions) is used. 

To demonstrate the ability of DVIB latent representations in dissecting shared and latent representations, we evaluate the quality of shared representations ($z_x^{(s)}$ and $z_x^{(p)}$) by predicting the shared labels (digit identities) and the quality of private representations by predicting the view-specific labels (rotation angles for $z_x^{(p)}$ and flip type for $z_y^{(p)}$, respectively) using simple linear classifier.

\begin{figure}
\centering
\includegraphics[width=0.9\textwidth]{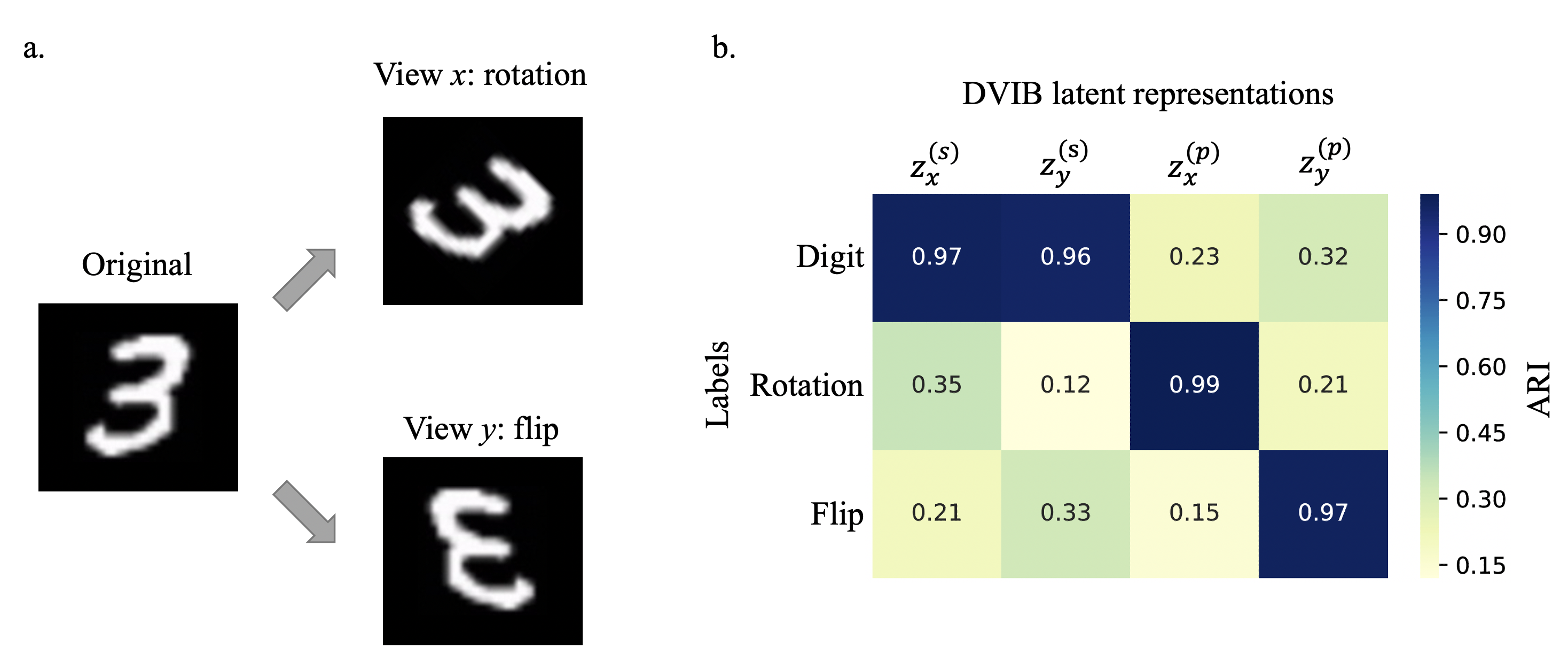}
\caption{(a) Example of simulated two-view data from original MNIST sample.  (b) Performance of DVIB latent representations in predicting view-specific and -shared labels. Adjusted Rand index (ARI) is calculated between predicted labels and groundtruths.} \label{re_1}
\end{figure}

This experiment gives an initial demonstration how DVIB can capture the shared and unique information in views. The results (Fig. \ref{re_1}) demonstrate the shared representations ($z_x^{(s)}$ and $z_x^{(p)}$) can best capture the shared label information (digit) while each private representation ($z_x^{(p)}$ and $z_x^{(p)}$) has higher accuracy in predicting each view-specific labels. Meanwhile, the private representation from one view has poor performance to infer the private label in the other view or the shared digit labels, indicating view-specific information from raw feature is isolated to the private representations as expected. 

\subsection{Evaluation using corrupted samples on ImageNet}

One important application for multiview learning is to compensate from the additional view and recover the corrupted information in either view. Here, we consider the classification problem on ImageNet \cite{deng2009imagenet}. To simulate the image corruptions, we add Gaussian noise for the generation of view $x$ and use defocus blur for the view $y$ \cite{michaelis2019dragon}.  We use the existing image corruption implementation \footnote{https://github.com/bethgelab/imagecorruptions} in the experiments. 
Again, to facilitate the efficient learning, we also employ the pretrained Inception-v3 framework to construct raw features. And multiview learning is built upon the deep features. 

To calibrate the performance of our method, we consider general multiview deep neural network (DNN) \cite{DBLP:conf/icml/NgiamKKNLN11}; multiview non-negative matrix factorization (M-NMF); deep canonical correlated autoencoder (DCCAE)\cite{pmlr-v37-wangb15} and its variational version (VCCA) \cite{wang2016deep}; variational autoencoder (VAE) \cite{kingma2013auto}; information theory based method deep variational information bottleneck (VIB) \cite{alemi2016deep}, multiview information bottleneck (MIB) \cite{federici2020learning}.
As we focus on the ability to remove the unwanted corruption information, we make use of the shared representations learned from two views. We note not all methods were designed for multiview study and we simply concatenate features from two views for these single-view methods. The classification  is performed by softmax regression on the latent representations learned by each method.

\begin{table}
\centering
\caption{Classification accuracy of ImageNet using joint latent representations learned from different method.}
\begin{tabular}{c|l|c}
\toprule
Method &  Description & Accuracy\\
\toprule
Baseline &  Simple feature concatenations.& 0.714\\
\hline
DNN &  Multiview neural network. & 0.763\\
\hline
M-NMF &  Joint matrix factorization. & 0.632\\
\hline
DCCAE &  A deep version of CCA. & 0.716\\
VCCA &  A variational version of CCA. & 0.758\\
\hline
VAE &  Variational autoencoder. & 0.782\\
\hline
VIB &  Variational information bottleneck. & 0.794\\
MIB &  Multiview information bottleneck. & 0.821\\
\hline
DVIB &  The proposed method. & \bf{0.852}\\
\bottomrule
\end{tabular}\label{tab_imagenet}
\end{table}

From the classification accuracies (Table \ref{tab_imagenet}), general multiview methods showed improved performance compared with feature concatenations. Variational methods (VCCA, VAE and VIB) have better robustness to the corruption than traditional methods DNN and NMF due to the design of variational inference. The recently proposed multiview information bottleneck method MIB, which  aims to learn informative  shared representations, further improved the accuracy. Our method, by explicitly formulating the property of shared latents from mutual information constrain, obtains the best accuracy.

\subsection{Robustness to image corruption levels}

Based on the results on ImageNet, here we ask how robust is each method's performance to the corruption levels of inputs. To investigate this, we consider to increase the corruption level for each single modality or for both, and evaluate the accuracies correspondingly.  As comparisons, we select top 5 methods from Table \ref{tab_imagenet} and the baseline.

\begin{figure}
\centering
\includegraphics[width=1\textwidth]{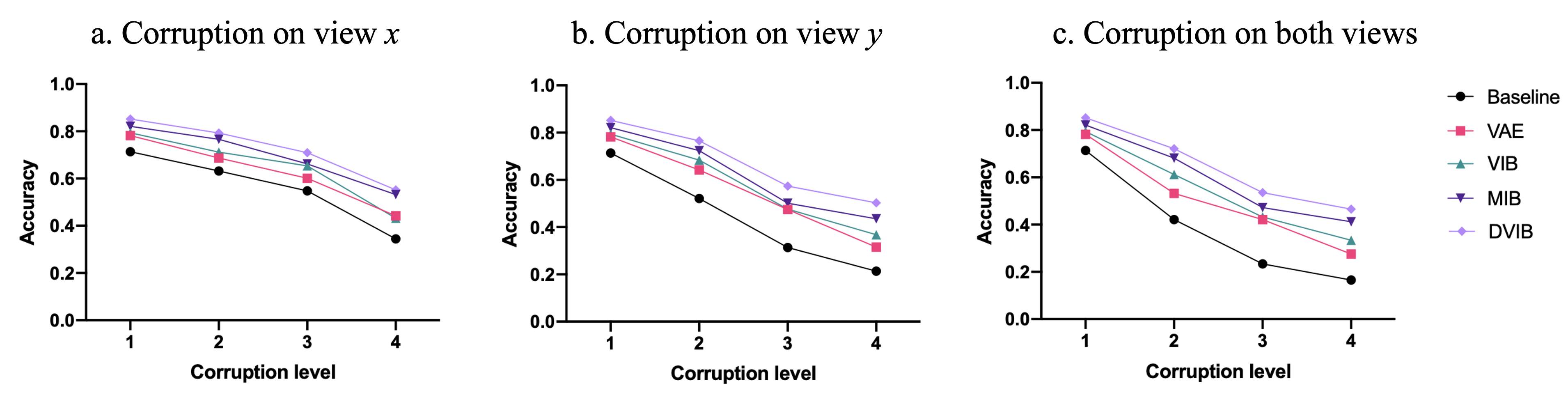}
\caption{Performance of classification on ImageNet with increasing corruption levels on (a) view $x$ alone, (b) view $y$ alone and (c) both.} \label{fig_corrupt}
\end{figure}

From the overall results (Fig. \ref{fig_corrupt}), corruptions in image blur level (view $y$) have stronger effects to the accuracy  than Gaussian noise (view $x$). All methods show decreasing trends when increasing the corruptions. MIB and DVIB maintain better performances compared with other method and DVIB has the best performance. It demonstrates the design of mutual information constrained structures improves the effective information extraction.

\section{Discussion}

In this work, we formulate the information disentanglement task from multiview data in the framework of information bottleneck. We explicitly define the desired information property of view-private and view-share latent representations with mutual information constrains and decompose each view. By deriving the variational bounds of mutual information, we effectively optimize the target function. We demonstrate the learned latent representations well preserve the view-shared and view-specific information and improve the classification robustness.

The DVIB model includes two hyperparameters $\lambda$ and $\beta$. How to simultaneously determine the appropriate values of these parameters is a challenge and requires further exploration. In our implementation, we consider same weights for two views as they are generated from the same source. However, for other types of multiview data which include different modalities (e.g. image and audio), this assumption might not stand. In the optimization of upper bound of mutual information (Eq. \ref{eq_up}), we omit the first term in final loss function with the assumption that neural network encoders for two view will defaultly produce difference encodings due to the random mapping property  neural network. However, it  requires abundant experiments study and rigorous mathematical proof. Taken together, with further extensive evaluation of the method, DVIB can be a potential powerful tool for the disentanglement of multiview data.

%
%
%
\bibliographystyle{unsrt}
\bibliography{mybibliography}

\newpage
\section*{Appendix}
\appendix
\addcontentsline{toc}{chapter}{Appendix}

\section{Lower bound for $I(x,z_x^{(s)})$}\label{app1}

\begin{align}  
  \begin{aligned}[t]
  I(x,z_x^{(s)}) 	&= \int{\int{p(x,z_x^{(s)})\log{\frac{p(x|z_x^{(s)})}{p(x)}}}}\,dx\,dz_x^{(s)}\\
      				&= \int{\int{p(x,z_x^{(s)})\log{\frac{p(x|z_x^{(s)})}{p(x)}\frac{q(x|z_x^{(s)})}{q(x|z_x^{(s)})}}}}\,dx\,dz_x^{(s)}\\
      				&= \int{\int{p(x,z_x^{(s)})\log{\frac{q(x|z_x^{(s)})}{p(x)}}}}\,dx\,dz_x^{(s)}+\int{\int{p(x,z_x^{(s)})\log{\frac{p(x|z_x^{(s)})}{q(x|z_x^{(s)})}}}}\,dx\,dz_x^{(s)}\\
      				&= \int{\int{p(x,z_x^{(s)})\log{\frac{q(x|z_x^{(s)})}{p(x)}}}}\,dx\,dz_x^{(s)}+\int{p(z_x^{(s)})\int{p(x|z_x^{(s)})\log{\frac{p(x|z_x^{(s)})}{q(x|z_x^{(s)})}}}}\,dx\,dz_x^{(s)}\\
      				&= \int{\int{p(x,z_x^{(s)})\log{\frac{q(x|z_x^{(s)})}{p(x)}}}}\,dx\,dz_x^{(s)}+\int{p(z_x^{(s)})KL[p(x|z_x^{(s)})||q(x|z_x^{(s)})]}\,dz_x^{(s)}\\
      				&\ge \int{\int{p(x,z_x^{(s)})\log{\frac{q(x|z_x^{(s)})}{p(x)}}}}\,dx\,dz_x^{(s)}\\
      				&= \int{\int{p(x,z_x^{(s)})\log{q(x|z_x^{(s)})}}}\,dx\,dz_x^{(s)}-\int{\int{p(x,z_x^{(s)})\log{p(x)}}}\,dx\,dz_x^{(s)}\\
      				&= \int{\int{p(x,z_x^{(s)})\log{q(x|z_x^{(s)})}}}\,dx\,dz_x^{(s)} +H(X)
  \end{aligned}
\end{align}
where $q(x|z_x^{(s)})$ is the approximation of $p(x|z_x^{(s)})$.
Similarly we can derive the lower bounds for $I(x,z_x^{(p)})$, $I(y,z_y^{(s)})$ and $I(y,z_y^{(p)})$.

\section{Lower bound for $I(y,z_x^{(s)})$}\label{app2}

$z_x^{(s)}$ is the shared latent representation generated from $x$. Following the reference \cite{federici2020learning}, we firstly enumerate several mutual information properties that we construct our bound upon:
\begin{align}  
  \begin{aligned}[t]
  I(\cdot;\cdot)  &\ge 0\\
  I(xy;z)&=I(y;z)+I(x;z|y)
  \end{aligned}
\end{align}

Then we have 

\begin{align}  
  \begin{aligned}[t]
  I(z_x^{(s)}, y)  & = I(z_x^{(s)};z_y^{(s)}y) - I(z_x^{(s)}; z_y^{(s)}|y)\\
  &=I(z_x^{(s)};z_y^{(s)}y)\\
  &=I(z_x^{(s)};z_y^{(s)})+I(z_x^{(s)};y|z_y^{(s)})\\
  &\ge I(z_x^{(s)};z_y^{(s)})
  \end{aligned}
\end{align}

 Here we assume the $z_y^{(s)}$ is a sufficient representation of $y$ so that $I(z_x^{(s)}; z_y^{(s)}|y)=0$.
 Again, we can follow the same formulation and derive the lower bounds for $I(x,z_y^{(s)})$.

\begin{align}  
  \begin{aligned}[t]
  I(z_y^{(s)}, x)    &\ge I(z_x^{(s)};z_y^{(s)})
  \end{aligned}
\end{align}

\begin{figure}
\centering
\includegraphics[width=1\textwidth]{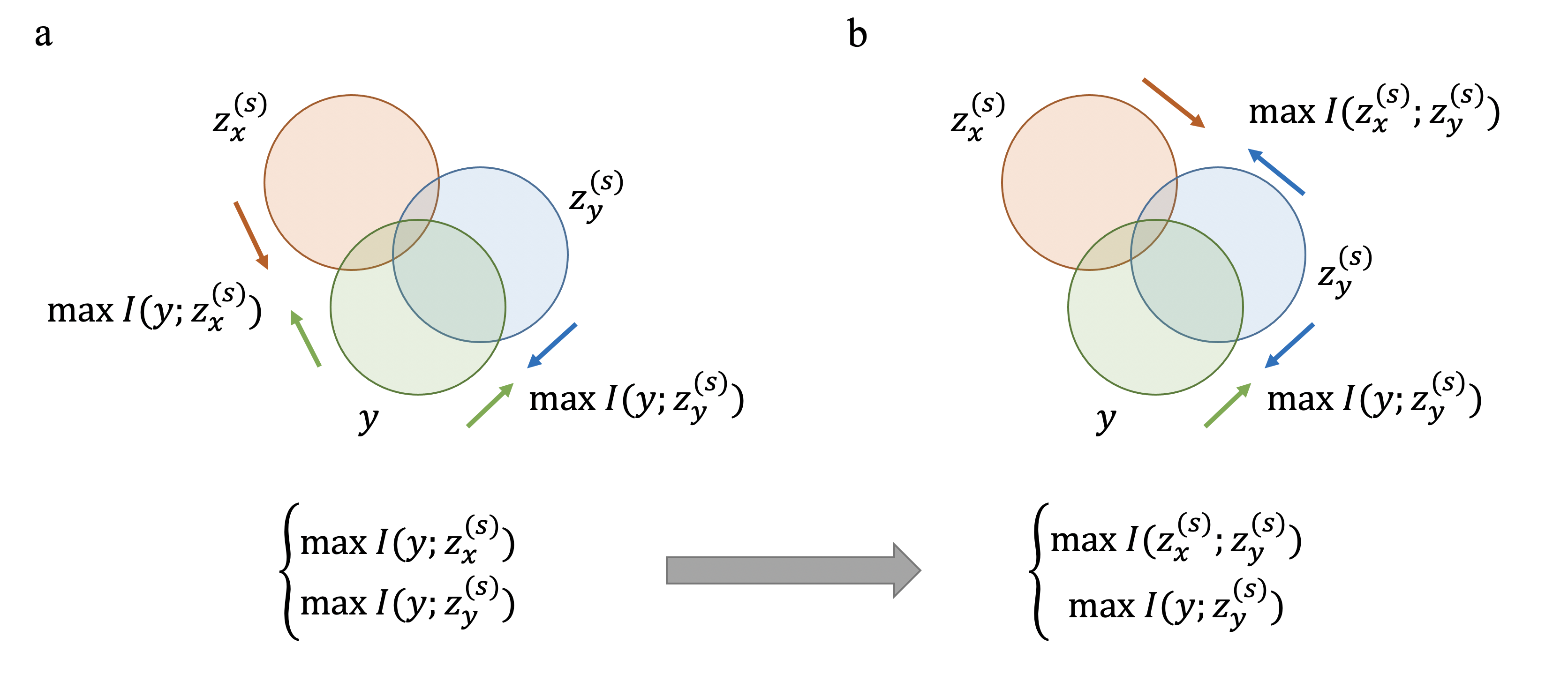}
\captionsetup{labelformat=empty}
\caption{Appendix Fig. 1 A demonstration of (a) original optimization target and (b) lower bound target.} 
\end{figure}

The above figure gives an illustration how the alternative optimization target approximate the original target using Venn diagram. 

\section{Upper bound for $I(y,z_x^{(p)})$}\label{app3}

By replacing the prior distribution $p(z_x^{(p)})$ with approximated marginal distribution $r(z_x^{(p)})$, we derive the upper bound of $I(y,z_x^{(p)})$: 

\begin{align}\label{upb}
  \begin{aligned}[t]
  I(y,z_x^{(p)}) 	&= \int{\int{p(y,z_x^{(p)})\log{\frac{p_x^{(p)}(z_x^{(p)}|y)}{p(z_x^{(p)})}}}}\,dy\,dz_x^{(p)}\\
  &= \int{\int{p(y,z_x^{(p)})\log{\frac{p_x^{(p)}(z_x^{(p)}|y)}{r(z_x^{(p)})}\frac{r(z_x^{(p)})}{p(z_x^{(p)})}}}}\,dy\,dz_x^{(p)}\\
  &= \int{\int{p(y,z_x^{(p)})\log{\frac{p_x^{(p)}(z_x^{(p)}|y)}{r(z_x^{(p)})}}}}\,dy\,dz_x^{(p)} -  \int{p(z_x^{(p)})\log{\frac{p(z_x^{(p)})}{r(z_x^{(p)})}}}\,dz_x^{(p)}\\
  &=\int{\int{p(y,z_x^{(p)})\log{\frac{p_x^{(p)}(z_x^{(p)}|y)}{r(z_x^{(p)})}}}}\,dy\,dz_x^{(p)}  - KL[p(z_x^{(p)})||r(z_x^{(p)})]\\
  &\le \int{\int{p(y,z_x^{(p)})\log{\frac{p_x^{(p)}(z_x^{(p)}|y)}{r(z_x^{(p)})}}}}\,dy\,dz_x^{(p)}\\
  &= \int{\int{p(y,z_x^{(p)})\log{\frac{p_x^{(p)}(z_x^{(p)}|y)}{p_y^{(p)}(z_y^{(p)}|y)}\frac{p_y^{(p)}(z_y^{(p)}|y)}{r(z_x^{(p)})}}}}\,dy\,dz_x^{(p)}\\
  &= \int{\int{p(y)KL[p_x^{(p)}(z_x^{(p)}|y)||p_y^{(p)}(z_y^{(p)}|y)]}}\,dy\,dz_x^{(p)} + \int{\int{p(y,z_x^{(p)})\log{\frac{p_y^{(p)}(z_y^{(p)}|y)}{r(z_x^{(p)})}}}}\,dy\,dz_x^{(p)}\\
  \end{aligned}
\end{align}
where $p_x^{(p)}$ and $p_y^{(p)}$ represent private encoders that learn latent representations $z_x^{(p)}, z_y^{(p)}$ from raw features $x, y$. Minimizing the lower bound $\int{\int{p(y,z_x^{(p)})\log{\frac{p_x^{(p)}(z_x^{(p)}|y)}{r(z_x^{(p)})}}}}\,dy\,dz_x^{(p)}$, it requires to   enlarge the distribution differences between  representations using the private encoder for view $x$ with input $y$ and the prior distribution of $z_x^{(p)}$. To better interpret this term,  we further expand the equation by add private encoder from $y$ and derive the final two-term upper bound (last line of Eq. \ref{upb}). The first term considers the KL-divergence between $p_x^{(p)}(z_x^{(p)}|y)$ and $p_y^{(p)}(z_y^{(p)}|y)$. To minimize this term, private encoders $p_x^{(p)}$ and $p_y^{(p)}$ need to produce different encoding with the same input $y$. The second term is to calculate the differences between $p_y^{(p)}(z_y^{(p)}|y)$ and $r(z_x^{(p)})$, which requires the encoding from one view should be different from the prior distribution of the other view. In our work, we use stochastic encoder to map the latent representation. Therefore, we assume the first term was in relatively small value and only optimize the second term in the loss function:

\begin{align}
  \begin{aligned}[t]
  \min{I(y,z_x^{(p)})}  &\equiv  \min{\int{\int{p(y,z_x^{(p)})\log{\frac{p_y^{(p)}(z_y^{(p)}|y)}{r(z_x^{(p)})}}}}\,dy\,dz_x^{(p)}}\\
  \end{aligned}
\end{align}

Symmetrically, we also have the approximation minimization target for $I(x,z_y^{(p)})$ as:

\begin{align}
  \begin{aligned}[t]
  \min{I(x,z_y^{(p)})}  &\equiv  \min{\int{\int{p(x,z_y^{(p)})\log{\frac{p_x^{(p)}(z_x^{(p)}|x)}{r(z_y^{(p)})}}}}\,dx\,dz_y^{(p)}}\\
  \end{aligned}
\end{align}

\section{Overall optimization target}\label{app4}

As shown in the main text, overall optimization function is given by:

\begin{align}  
  \begin{aligned}[t]
  \max ~I_{total}&=I_x+I_y\\ 
  &\ge I_{LB}(x;z_x^{(s)};z_x^{(p)})+ I_{LB}(y;z_y^{(s)};z_y^{(p)}) + \\
  &~~~~\lambda I(z_x^{(s)},z_y^{(s)})  - \beta I^{(c)}(z_x^{(p)},z_y^{(p)})\\
  \end{aligned}
\end{align}

where 

\begin{align}  
  \begin{aligned}[t]
  I_{LB}(x;z_x^{(s)};z_x^{(p)})&=\mathbb{E}_{p(x,z_x^{(s)})}\log{q(x|z_x^{(s)})} + \mathbb{E}_{p(x,z_x^{(p)})}\log q(x|z_x^{(p)})\\
  I_{LB}(y;z_y^{(s)};z_y^{(p)})&=\mathbb{E}_{p(y,z_y^{(s)})}\log q(y|z_y^{(s)}) + \mathbb{E}_{p(y,z_y^{(p)})}\log q(y|z_y^{(p)})\\
  I^{(c)}(z_x^{(p)},z_y^{(p)})&\stackrel{{\beta_x =\beta_y}}{=}\mathbb{E}_{p(y,z_x^{(p)})}\log{\frac{p_y^{(p)}(z_y^{(p)}|y)}{r(z_x^{(p)})}}+\\
  &~~~~~~~~~\mathbb{E}_{p(x,z_y^{(p)})}\log{\frac{p_x^{(p)}(z_x^{(p)}|x)}{r(z_y^{(p)})}}\\
  \end{aligned}
\end{align}
 $I_{LB}(x;z_x^{(s)};z_x^{(p)})$ and $I_{LB}(y;z_y^{(s)};z_y^{(p)})$ are lower bounds of $I(x;z_x^{(s)};z_x^{(p)})$ and $I(y;z_y^{(s)};z_y^{(p)}$; $I^{(c)}(z_x^{(p)},z_y^{(p)})$ represents the cross mutual information between private representations of two views. As two  latent representations can be defined in the same prior distribution, we can simplify the weight of this term as $\beta=\beta_x =\beta_y$. $I(z_x^{(s)},z_y^{(s)})$ is the  mutual information between  shared representations from two views. We also assume the $\lambda = \lambda_x =\lambda_y$ for the same reason.

 \end{document}